\documentclass{article}
\usepackage{ijcai19}

\usepackage{times}
\usepackage{soul}
\usepackage{url}
\usepackage[utf8]{inputenc}
\usepackage[small]{caption}
\usepackage{graphicx}
\usepackage{amsmath}
\usepackage{booktabs}
\usepackage{algorithm}
\usepackage{algorithmic}
\usepackage{color}
\usepackage{comment}
\usepackage{amssymb,amsmath}
\usepackage{multirow}
\usepackage{subfigure}
\title{Multi-View Multiple Clustering}
\author{
Shixing Yao$^1$\and
Guoxian Yu$^1$\and
Jun Wang$^{1}$\and
Carlotta Domeniconi$^2$\and
Xiangliang Zhang$^{3}$
\affiliations
$^1$College of Computer and Information Sciences, Southwest University, Chongqing, China\\
$^2$Department of Computer Science, George Mason University, VA, USA\\
$^3$CEMSE, King Abdullah University of Science and Technology, Thuwal, SA\\
$^4$Fourth Affiliation\\
\emails
\{ysx, gxyu, kingjun\}@swu.edu.cn,
carlotta@cs.gmu.edu,
xiangliang.zhang@kaust.edu.sa
}
\begin{document}
\maketitle
\begin{abstract}
Multiple clustering aims at exploring alternative clusterings to organize the data into meaningful groups from different perspectives. Existing multiple clustering algorithms are designed for {\bf single-view} data. We assume that the \emph{individuality} and \emph{commonality} of multi-view data can be leveraged to generate high-quality and diverse clusterings. To this end, we propose a novel multi-view multiple clustering (MVMC) algorithm. MVMC first adapts multi-view self-representation learning to explore the individuality encoding matrices and the shared commonality matrix of multi-view data. It additionally reduces the redundancy (i.e., enhancing the individuality) among the matrices using the Hilbert-Schmidt Independence Criterion (HSIC), and collects shared information by forcing the shared matrix to be smooth across all  views. It then uses matrix factorization on the individual matrices, along with the shared matrix, to generate diverse clusterings of high-quality. We further extend multiple co-clustering on multi-view data and propose a solution called multi-view multiple co-clustering (MVMCC). Our empirical study shows that MVMC (MVMCC) can exploit multi-view data to generate multiple high-quality and diverse clusterings (co-clusterings), with superior performance to the state-of-the-art methods.
\end{abstract}

\vspace{-3mm}
\section{Introduction}
The goal of clustering is to partition samples into disjoint groups to facilitate the discovery of hidden patterns in the data. Traditional clustering algorithms are designed for single-view data. With the diffusion of the internet of things and of big data, samples can be easily collected from different sources, or observed from different views. For example, a video can be characterized using image signals and audio signals, and a given news can be reported in different languages. Objects with diverse feature views are typically called multi-view data. It's recognized that the integration of information contained in multiple views can achieve consolidated data clustering \cite{chao2017mvcsurvey}. Many multi-view clustering methods have been developed to extract comprehensive information from multiple feature views; examples are co-training based  \cite{kumar2011co}, multiple kernel learning  \cite{gonen2011multiple}, and subspace learning based \cite{cao2015diversity,luo2018consistent}. However, the aforementioned clustering methods typically provide a \emph{single} clustering, which may fail to reveal the high-quality but all diverse alternative clusterings of the same data.

\begin{figure}[t]
  \centering
  \includegraphics[width=7cm,height=5cm]{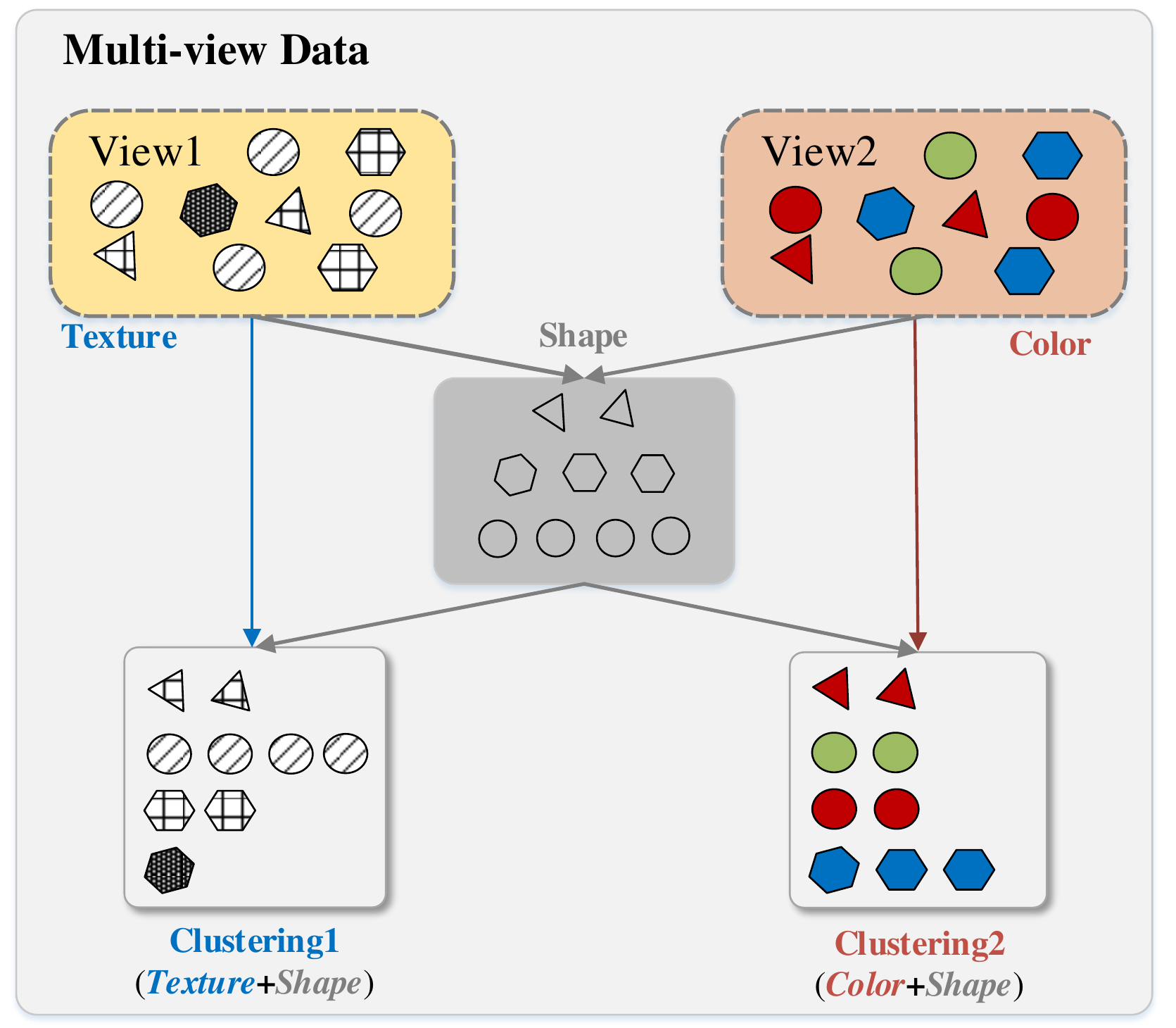}\\
  \caption{An example of multi-view multiple clustering. Two alternative clusterings (texture+shape and color+shape) can be generated using the \emph{commonality} (shape) and the \emph{individuality} (texture and color) information of the same multi-view objects. }
  \label{fig:case}
\end{figure}

For example, in Figure \ref{fig:case}, we have a collection of objects represented by a texture view and a color view. We can group the objects based on their shared shapes. By leveraging the {\bf commonality} and the {\bf individuality} of these multi-view objects, we can obtain two alternative clusterings (\emph{texture+shape} and \emph{color+shape}), as shown in the bottom of the figure.  From this example, we can see that multi-view data include not only the commonality information for generating high-quality clustering (as multi-view clustering does), but also the individual (or specific) information for generating diverse clusterings (as multiple clustering aims to achieve) \cite{bailey2013alternative}.

To explore different clusterings of the given data, multiple clustering has emerged as a new branch of clustering in recent years.  Some approaches seek diverse clusterings in alternative to those already explored, by enforcing the new ones to be different \cite{bae2006coala,davidson2008finding,yang2017non}; other solutions simultaneously seek multiple clusterings  by reducing their correlation \cite{caruana2006meta,jain2008simultaneous,dang2010generation,wang2018mcc}, or by seeking orthogonal (or independent) subspaces and clusterings therein \cite{niu2010multiple,ye2016generalized,mautz2018kdd,wang2019MISC}. However, these multiple clustering methods are designed for \emph{single-view} data.




Based on the discussed example in Figure \ref{fig:case}, we leverage the individuality and the commonality of multi-view data to generate high-quality and diverse clusterings, and we propose an approach called multi-view multiple clustering (MVMC) to achieve this goal. To the best of our knowledge, MVMC is the \emph{first attempt} to encompass both multiple clusterings and multi-view clustering, where the former focuses on generating diverse clusterings from a single data view, and the latter focuses on a single consensus clustering by summarizing the information from different views. MVMC first extends  multi-view self-representation learning \cite{luo2018consistent} to explore the individuality information encoding matrices and the commonality information matrix shared across views. To obtain more credible commonality information from multiple views, we force the commonality information matrix to be smooth across all views. In addition, we use the Hilbert-Schmidt Independence Criterion (HISC)\cite{gretton2005measuring} to enhance the individuality between matrices, and consequently increase the diversity between clusterings. We then use matrix factorization to jointly factorize each individuality matrix (for diversity) and the commonality matrix (for quality) into a clustering indicator matrix and a basis matrix. To simultaneously seek the individual and common data matrices, and diverse clusterings therein, we use an alternating optimization technique to solve the unified objective. In addition, we extend multiple co-clustering \cite{tokuda2017multiple,wang2018mcc} to the multi-view scenario, and term the extended approach as multi-view multiple co-clustering (MVMCC).

The main contributions of our work are summarized as follows:
\begin{itemize}
\item We study how to generate {\bf multiple clusterings from multi-view} data, which is an interesting and challenging problem, but largely overlooked. The problem we address is different from existing multi-view clustering, which generates a single clustering by leveraging multiple views, and also different from multiple clustering, which  produces alternative clusterings from single-view data.

\item We introduce a {\bf unified objective function} to simultaneously seek {\bf multiple individuality information} encoding matrices, and the {\bf commonality information} encoding matrix. This unified function can leverage  the individuality to generate  diverse clusterings, and the commonality to boost the quality of the generated clusterings. We further adopt an alternative optimization technique to solve the unified objective.

\item Extensive experimental results show that  MVMC (MVMCC) performs considerably better than existing multiple clustering (co-clustering) algorithms \cite{cui2007non,jain2008simultaneous,niu2010multiple,ye2016generalized,yang2017non,tokuda2017multiple,wang2018mcc,wang2019MISC} in exploring multiple clusterings and co-clusterings.

\end{itemize}

\vspace{-4mm}
\section{The Proposed Methods}
\subsection{Multi-View Multiple Clustering}
Suppose $\mathbf{X}^v \in \mathbb{R}^{d_v \times n}$ denotes the feature data matrix of the $v$-th view, $v\in \{ 1,2,\cdots,m \}$, for $n$ objects in the $d_v$ dimensional space. We aim to generate $h$ (provided by the user) different clusterings from $\{\mathbf{X}^v\}_{v=1}^m$ using the shared and individual information embedded in the data matrices. Most multi-view clustering approaches in essence focus on the shared and complementary information of multiple data views to generate a consolidated clustering \cite{chao2017mvcsurvey}.  By viewing each subspace as a feature view, an intuitive solution to explore multiple clusterings on multi-view data is to first concatenate different feature views, and then apply subspace-based multiple clustering methods on the concatenated feature vectors \cite{niu2010multiple,ye2016generalized,wang2019MISC}.


To find high-quality and diverse multiple clusterings, we should make concrete use of the individuality and commonality of multi-view data. The individuality helps to explore diverse clusterings, while the commonality coordinates the diverse clusterings to capture the common knowledge of multi-view data. To explore the individuality and commonality of multi-view data, we extend the multi-view self-representation learning \cite{cao2015diversity,luo2018consistent} as follows:
\begin{equation}
\label{Eq1}
\small
\centering
\begin{split}
    J_D(\{\mathbf{D}^k\}_{k=1}^h, \mathbf{U}) &= \sum_{k=1}^h \frac{1}{m} \sum_{v=1}^m \parallel \mathbf{X}^v- \mathbf{X}^v(\mathbf{U} +\mathbf{D}^k)\|_F^2\\ & +\lambda_1 \Phi_1(\{\mathbf{D}^k\}_{k=1}^h)+ \lambda_2 \Phi_2(\mathbf{U})
\end{split}
\end{equation}
where $\mathbf{U} \in \mathbb{R}^{n \times n}$ is specified to encode the commonality of the data matrices $\{\mathbf{X}^v\}_{v=1}^m$, and $\mathbf{D}^k \in \mathbb{R}^{n \times n}$ is used to encode the individuality of the $k$-th ($k \in \{1,2,,\cdots,h\}$) group of views. $\Phi_1(\{\mathbf{D}^k\}_{k=1}^h)$ and $\Phi_2(\mathbf{U})$ (defined later) are two constraints used to enhance the individuality and commonality. The multi-view self-representation learning in \cite{cao2015diversity,luo2018consistent} requires $h=m$. In contrast, Eq. (\ref{Eq1}) does not have this requirement. As a result, the  group-wise individuality and diversity are jointly considered, and the number of alternative clusterings can be adjusted by the user. The assumption for the linear representation in Eq. (\ref{Eq1}) is that a data sample can be expressed as a linear combination of other samples in the subspace. This assumption is widely-used in sparse representation \cite{wright2010sr} and low-rank representation learning \cite{liu2013llr}.

\cite{luo2018consistent} recently combined $\mathbf{U}$ and $\{\mathbf{D}^k\}_{k=1}^m$ into an integrated co-association matrix of samples, and then applied spectral clustering to seek a consistent clustering. Their empirical study shows that the individual information encoded by $\mathbf{D}^k$ helps  producing a robust clustering. However, since $\mathbf{X}^v$ and $\mathbf{X}^{v'}(v' \neq v)$  describe the same objects using different types of features, the matrix $\mathbf{D}^k$ resulting from Eq. (\ref{Eq1}) may still have a large information overlap with $\mathbf{D}^{k'}$. As a result, the expected individuality of  $\mathbf{D}^k$ and $\mathbf{D}^{k'}$ cannot be guaranteed. This overlapping of information is not necessarily an issue for multi-view clustering, which aims at finding a single clustering, but it is for our problem, where multiple clusterings of both high-quality and diversity are expected.

To enhance the diversity between individuality encoding (representation) matrices $\{\mathbf{D}^k\}_{k=1}^h$, we approximately quantify the diversity based on the dependency between these matrices. The smaller the dependency between data matrices is, the larger their diversity is, since the matrices are less correlated. Various measurements can be used to evaluate the dependence between variables. Here, we adopt the Hilbert-Schmidt Independence Criterion (HSIC) \cite{gretton2005measuring}, for its simplicity, solid theoretical foundation, and capability in measuring both linear and nonlinear dependence between variables.
HSIC computes the squared norm of the cross-covariance operator over $\mathbf{D}^k$ and $\mathbf{D}^{k'}$ in Hilbert Space to estimate the dependency. The empirical HISC does not have to explicitly compute the joint distribution of  $\mathbf{D}^k$ and $\mathbf{D}^{k'}$, it is give by:
\begin{equation}
\label{Eq2}
\small
    HSIC(\mathbf{D}^k,\mathbf{D}^{k'})= (n-1)^{-2} tr(\mathbf{K}^k \mathbf{H} \mathbf{K}^{k'} \mathbf{H})
\end{equation}
where $\mathbf{K}^k, \mathbf{K}^{k'}, \mathbf{H} \in \mathbb{R}^{n \times n}$, $\mathbf{K}^k$ and $\mathbf{K}^{k'}$ are  used to measure the kernel induced similarity between vectors of $\mathbf{D}^k$ and $\mathbf{D}^{k'}$, respectively. $\mathbf{H}=\delta_{ij}-1/n$, $\delta_{ij}=1$ if $i=j$, $\delta_{ij}=0$ otherwise. In this paper, we adopt the inner product kernel to specify $\mathbf{K}^k=(\mathbf{D}^k)^T \mathbf{D}^k$ $\forall k \in \{1,2,\cdots,h\}$. Then we minimize the overall HISC on $h$ individuality matrices to reduce the redundancy between them and specify $\Phi_1(\mathbf{D}^k)$ as follows:
\begin{equation}
\label{Eq3}
\centering
\small
\begin{split}
    \Phi_1(\{\mathbf{D}^k\}_{k=1}^h) &= \sum_{k = 1, k \neq k'}^{h} HSIC(\mathbf{D}^k, \mathbf{D}^{k'})\\
    &= \sum_{k = 1, k \neq k'}^{m} (n-1)^{-2} tr(\mathbf{D}^k \mathbf{H} \mathbf{K}^{k'} \mathbf{H} (\mathbf{D}^k)^T) \\
    &= \sum_{k=1}^h tr(\mathbf{D}^k \widetilde{\mathbf{K}}^k (\mathbf{D}^k)^T)
\end{split}
\end{equation}
where  $\widetilde{\mathbf{K}}^k=(n-1)^{-2}\sum_{k'=1, k'\neq k}^{m} \mathbf{H} \mathbf{K}^{k'} \mathbf{H}$.

Inspired by subspace-based multi-view learning \cite{gao2015multi,chao2017mvcsurvey} and manifold regularization \cite{belkin2006mr}, we specify $\Phi_2(\mathbf{U})$ in Eq. (\ref{Eq1}) to collect more shared information from multiple data views  as follows:
\begin{equation}
\label{Eq4}
\centering
\small
\Phi_2(\mathbf{U})=\sum_{v=1}^V \sum_{i,j=1}^n \parallel \mathbf{u}_i-\mathbf{u}_j \parallel_2^2 \mathbf{W}^v_{ij}=tr(\mathbf{U} \widetilde{\mathbf{L}} \mathbf{U}^T)
\end{equation}
where $\mathbf{W}^v_{ij}$ is the feature similarity between  $\mathbf{x}_i^v$ and $\mathbf{x}_j^v$. To compute  $\mathbf{W}^v$, we simply adopt an $\epsilon=5$ nearest neighborhood graph  and use the Gaussian heat kernel (with kernel width as the standard deviation of the distance between samples) to quantify the similarity between neighborhood samples. $\widetilde{\mathbf{L}}=\sum_{v=1}^m (\mathbf{\Lambda}^v-\mathbf{W}^v)$ and $\mathbf{\Lambda}^v$ is a diagonal matrix with $\mathbf{\Lambda}^v_{ii}=\sum_{j=1}^n \mathbf{W}^v_{ij}$. Minimizing $\Phi_2(\mathbf{U})$ can guide $\mathbf{U}$ to encode consistent and complementary information shared across views.  In this way,  the quality of diverse clusterings can be boosted using enhanced shared information.

Given the equivalence between matrix factorization based clustering and spectral clustering  (or $k$-means clustering), we adopt the widely used  semi-nonnegative matrix factorization \cite{ding2010convex} to explore the $k$-th clustering on $\mathbf{U}+\mathbf{D}^k$ as follows:
\begin{equation}
\label{Eq6}
\small
\mathbf{U}+\mathbf{D}^k=\mathbf{B}^k(\mathbf{R}^k)^T
\end{equation}
where $\mathbf{R}^k \in \mathbb{R}^{n \times r_k}$ and $\mathbf{B}^k \in \mathbb{R}^{n \times r_k}$ ($r_k$ is the number of sample clusters)  are the clustering indicator matrix and the basis matrix, respectively. Here, the $k$-th clustering is generated not only with respect to $\mathbf{D}^k$, but also to $\mathbf{U}$, which encodes the shared information of multi-view data. As such, the explored $k$-th clustering (encoded by $\mathbf{R}^k$)   not only reflects the individuality of views in the $k$-th group, but also captures the commonality of all data views. As a consequence, a high-quality, and yet diverse clustering can be generated.

The above process first explores the individual information matrices and the shared information matrix, and then generates diverse clusterings on the data matrices. A sub-optimal solution may be obtained as a result because the two steps are performed separately. To avoid this, we advocate to simultaneously optimize $\{\mathbf{D}^k\}_{k=1}^h$ and the diverse clusterings $\{\mathbf{R}^k\}_{k=1}^h$ therein, and formulate a unified objective function for MVMC as follows:
\begin{equation}
\label{Eq7}
\begin{split}
\small
&J_{MC}(\{\mathbf{D}^k\}_{k=1}^h, \{\mathbf{R}^k\}_{k=1}^h, \mathbf{U}) \\
&= \frac{1}{h}\sum_{k=1}^h \parallel (\mathbf{U} + \mathbf{D}^k) - \mathbf{B}^k(\mathbf{R}^k)^T \parallel_F^2\\
&+ \lambda_1 tr(\mathbf{D}^k \widetilde{\mathbf{K}}^k (\mathbf{D}^k)^T) + \lambda_2 tr(\mathbf{U} \widetilde{\mathbf{L}} \mathbf{U}^T) \\
&s.t. \  \mathbf{X}^v = \mathbf{X}^v(\mathbf{U} + \mathbf{D}^k), \ v \in \{1,2, \cdots,m\}
 \end{split}
\end{equation}
By solving Eq. (\ref{Eq7}), we can simultaneously obtain multiple diverse clusterings of quality by leveraging the commonality and individuality information of multiple views. Our experiments confirm that MVMC can generate multiple clusterings with enhanced diversity and quality. In addition, it outperforms the counterpart algorithms \cite{cui2007non,jain2008simultaneous,niu2010multiple,ye2016generalized,wang2019MISC}, which concatenate multiple data views into a composite view, and then explore multiple clusterings in the subspaces of the composite view.

Binary matrices $\{\mathbf{R}^k\}_{k=1}^h$ are hard to directly optimize. As such, we relax the entries of $\{\mathbf{R}^k\}_{k=1}^h$ to nonnegative numeric values. Since Eq. (\ref{Eq7}) is not jointly convex for $\{\mathbf{D}^k\}_{k=1}^h$, $\mathbf{U}$ and $\{\mathbf{R}^k\}_{k=1}^h$, it is
unrealistic to find the global optimal values for all the variables. Here, we solve Eq. (\ref{Eq7}) via the alternating optimization method, which
alternatively optimizes one variable, while fixing the other variables. The detailed optimization process can be viewed in the supplementary file due to the limitation of space.

\vspace{-2mm}
\subsection{Multiple Views Multiple Co-Clusterings}
Multiple co-clustering algorithms recently have also been suggested to explore alternative co-clusterings from the same data \cite{tokuda2017multiple,wang2018mcc}. Multiple co-clustering methods aim at exploring multiple two-way clusterings, where both samples are features are clustered. In contrast, multiple clustering techniques only explore diverse one-way clusterings, where only samples (or only features) are clustered. Based on the merits of matrix tri-factorization in exploring co-clusters \cite{wang2011fast,wang2018mcc}, we can seek
multiple co-clusterings on multiple views by optimizing an objective function as follows:
\begin{equation}
\label{Eq8}
\begin{split}
\small
&J_{MCC}(\{\mathbf{D}^v\}_{v=1}^m, \{\mathbf{R}^v\}_{v=1}^m, \{\mathbf{C}^v\}_{v=1}^m, \mathbf{U}) \\
&= \frac{1}{m}\sum_{v=1}^m \parallel \mathbf{X}^v(\mathbf{U} + \mathbf{D}^v) - \mathbf{C}^v  \mathbf{S}^v (\mathbf{R}^v)^T  \parallel_F^2\\
&+ \lambda_1 tr(\mathbf{D}^v \mathbf{K} (\mathbf{D}^v)^T) + \lambda_2 tr(\mathbf{U} \widetilde{\mathbf{L}} \mathbf{U}^T)  \\
&s.t. \  \mathbf{X}^v = \mathbf{X}^v(\mathbf{U} + \mathbf{D}^v), \mathbf{C}^v \ge 0, \ \mathbf{R}^v \ge 0
\end{split}
\end{equation}
where  $\mathbf{C}^v \in \mathbb{R}^{d_v \times c_v}$ and $\mathbf{R}^v \in \mathbb{R}^{n \times r_k}$  correspond to the row-cluster (grouping features) and column-cluster (grouping samples) indicator matrices of the $h$-th co-clustering. $\mathbf{S}^{k} \in \mathbb{R}^{c_v \times r_v}$ plays the role of absorbing the different scaling factors of $\mathbf{R}^v$ and $\mathbf{C}^v$ to minimize the squared error. Here we fix $h=m$ for MVMCC, since different feature views have different numbers of features.  Eq. (\ref{Eq8}) can be  optimized following the similar procedure for optimizing Eq. (\ref{Eq7}), which is provided in the supplementary file due to page limit.

\vspace{-2mm}
\section{Experimental Results and Analysis}
\subsection{Experimental Setup}
In this section, we evaluate our proposed MVMC and MVMCC on five widely-used multi-view datasets \cite{li2015large,tao2018reliable}, which are described in  Table \ref{table1}. The datasets have different number of views and are from different domains. \textbf{Caltech-7}\footnote{\url{https://github.com/yeqinglee/mvdata}} and \textbf{Caltech-20} \cite{li2015large} are two subsets of Caltech-101, which contains only 7 and 20 classes, respectively. The creation of these subsets is due to the unbalance of the number of data in each class of Caltech-101. Each sample is made of 6 views on the same image. \textbf{Mul-fea digits}\footnote{\url{https://archive.ics.uci.edu/ml/datasets/Multiple+Features}} is comprised of 2,000 data points from 0 to 9 digit classes, with 200 data points for each class. There are six public features available: 76 Fourier coefficients of the character shapes, 216 profile correlations, 64 Karhunenlove coefficients, 240 pixel averages in $2 \times 3$ windows, 47 Zernike moments, and 6 morphological features. \textbf{Wiki article}\footnote{\url{http://www.svcl.ucsd.edu/projects/crossmodal/}} contains selected sections from the Wikipedia's featured articles collection. We considered only the 10 most populated categories. It contains two views: text and image. \textbf{Corel}\footnote{\url{http://www.cais.ntu.edu.sg/˜chhoi/SVMBMAL/}} \cite{tao2018reliable} consists of 5000 images from 50 different categories. Each category has 100 images. The features are color histogram (9), edge direction histogram (18), and WT (9). \textbf{Mirflickr}\footnote{\url{http://press.liacs.nl/mirflickr/mirdownload.html}} contains 25,000 instances collected from Flicker. Each instance consists of an image and its associated textual tags. To avoid noise, here we remove textual tags that appear less than 20 times in the dataset, and then delete instances without textual tags or semantic labels. This process gives us 16,738 instances.

\begin{table}[h!tbp]
\scriptsize
		\caption{Characteristics of multi-view datasets. $n$ is the number of samples, $d_v$ is the dimensionality of samples, 'classes' is the number of ground-truth clusters, and $m$ is the number of views.}
		\centering
\begin{tabular}{c|c|c|c|c}
	\hline
	Datasets &$n$ &$d_v$ &classes &$m$\\
	\hline
    Caltech-7&1474&[40,48,254,1984,512,928]&7&6\\
	Caltech-20&2386&[40,48,254,1984,512,928]&20&6\\
	Mul-fea digits&2000&[76,216,64,240,47,6]&10&6\\
    Wiki article&2866&[128,10]&10&2\\
    Corel&5000&[9,18,9]&50&3\\
    Mirflickr&16738&[150,500]&24&2\\
	\hline
\end{tabular}\label{table1}
\end{table}

Multiple clusterings need to quantify the quality and diversity of alternative clusterings.
To measure quality, we use the widely-adopted Silhouette Coefficient (SC) and Dunn Index (DI) as the internal index.
 \emph{Large} values of SC and DI indicate a \emph{high} quality clustering.
To quantify the redundancy between alternative clusterings, we use Normalized Mutual Information(NMI) and Jaccard Coefficient(JC) as  external indices.
  \emph{Smaller}   values of NMI and JC indicate   \emph{smaller}   redundancy between alternative clusterings. All these metrics have been used in the multiple clustering literature \cite{bailey2013alternative}. The formal definitions of these metrics, omitted here to save space, can be found in \cite{bailey2013alternative,yang2017non}.

\vspace{-2mm}

\subsection{Discovering multiple one-way clusterings and multiple co-clusterings}
We compare the one-way multiple clusterings found by MVMC against Dec-kmeans \cite{jain2008simultaneous}, MNMF \cite{yang2017non}, OSC \cite{cui2007non}, mSC \cite{niu2010multiple}, ISAAC \cite{ye2016generalized}, and MISC \cite{wang2019MISC}. We also compare the multiple co-clusterings found by MVMCC against MultiCC \cite{wang2018mcc} and MCC-NBMM \cite{tokuda2017multiple}.  The input parameters of the comparing methods are set as the authors suggested in their papers or shared code. The parameter values of MVMC and MVMCC are  $\lambda_1=10$, $\lambda_2=100$ and $h = 2$ for multiple one-way clusterings, and $h=m$ for multiple co-clusterings. None of existing multiple clusterings algorithms can work on multiple view data, we concatenate the feature vectors of multiple view data and then run these comparing methods on the concatenated vectors to seek alternative clusterings. Our MVMC and MCMCC directly run on the multiple view data, without such feature concatenation.

 We use $k$-means to generate the reference clustering for MNMF, and then use their respective solutions to generate two alternative clustering ($\mathcal{C}_1$, $\mathcal{C}_2$).
We downloaded the source code of MNMF, ISAAC, MultiCC, MISC, and MCC-NBMM, and implemented the other methods (Dec-kmeans, mSC, and OSC) following the respective original papers. Input parameters of the comparing methods  were fixed or optimized as suggested by the authors. Following the experimental protocol adopted by these methods, we quantify the average clustering quality on $\mathcal{C}_1$ and $\mathcal{C}_2$, and measure the diversity between $\mathcal{C}_1$ and $\mathcal{C}_2$. We fix the number of row-clusters $r_k$ for each clustering as the respective number of classes of each dataset, as listed in Table \ref{table1}. For co-clustering, we adopt a widely used technique \cite{monti2003consensus} to determine the number of column clusters $c_k$. Detailed parameter values can be viewed in the supplementary file.

\begin{table*}[h!tbp]
\scriptsize
	\caption{Quality and Diversity of the various competing methods on finding multiple clusterings. $\uparrow$($\downarrow$) indicates the direction of preferred values for the corresponding measure. $\bullet / \circ$ indicates whether MVMC is statistically (according to pairwise $t$-test at 95\% significance level) superior/inferior to the other method.}
	\resizebox{170mm}{28mm}{
	\begin{tabular}{c| c| c c c c c c c}
		\hline
		& & Dec-kmeans &ISAAC &MISC &MNMF &mSC &OSC &MVMC\\
		\hline
        \multirow{4}[2]{*}{Caltech-7}
        &SC$\uparrow$
        & 0.049$\pm$0.002$\bullet$ & 0.235$\pm$0.011$\circ$ & 0.201$\pm$0.002$\circ$ & 0.234$\pm$0.001$\circ$ & 0.163$\pm$0.008$\circ$ & 0.261$\pm$0.004$\circ$ & 0.140$\pm$0.002 \\
        &DI$\uparrow$
        & 0.044$\pm$0.000$\bullet$ & 0.034$\pm$0.001$\bullet$ & 0.048$\pm$0.000$\bullet$ & 0.034$\pm$0.000$\bullet$ & 0.056$\pm$0.000$\bullet$ & 0.066$\pm$0.000 & 0.062$\pm$0.000 \\
        &NMI$\downarrow$
        & 0.024$\pm$0.000$\bullet$ & 0.485$\pm$0.023$\bullet$ & 0.513$\pm$0.016$\bullet$ & 0.022$\pm$0.000$\bullet$ & 0.152$\pm$0.002$\bullet$ & 0.693$\pm$0.015$\bullet$ & 0.006 $\pm$ 0.000 \\
        &JC$\downarrow$
        & 0.126$\pm$0.001$\bullet$ & 0.363$\pm$0.008$\bullet$ & 0.349$\pm$0.001$\bullet$ & 0.094$\pm$0.000$\bullet$ & 0.136$\pm$0.001$\bullet$ & 0.522$\pm$0.046$\bullet$ & 0.076 $\pm$ 0.000 \\
        \hline
        \multirow{4}[2]{*}{Caltech-20}
        &SC$\uparrow$
        & -0.124 $\pm$ 0.001$\bullet$ & 0.085 $\pm$ 0.000$\circ$ & 0.036 $\pm$ 0.000$\circ$ & -0.169 $\pm$ 0.000$\bullet$ & -0.172 $\pm$ 0.006$\bullet$ & 0.196 $\pm$ 0.001$\circ$ & 0.004 $\pm$ 0.000 \\
        &DI$\uparrow$
        & 0.026 $\pm$ 0.000$\bullet$ & 0.035 $\pm$ 0.000$\bullet$ & 0.033 $\pm$ 0.000$\bullet$ & 0.009 $\pm$ 0.000$\bullet$ & 0.028 $\pm$ 0.000$\bullet$ & 0.056 $\pm$ 0.000$\bullet$ & 0.183 $\pm$ 0.000\\
        &NMI$\downarrow$
        & 0.056 $\pm$ 0.000$\bullet$ & 0.475 $\pm$ 0.011$\bullet$ & 0.489 $\pm$ 0.013$\bullet$ & 0.052 $\pm$ 0.002$\bullet$ & 0.240 $\pm$ 0.003$\bullet$ & 0.715 $\pm$ 0.025$\bullet$ & 0.027 $\pm$ 0.000 \\
        &JC$\downarrow$
        & 0.050 $\pm$ 0.001$\bullet$ & 0.222 $\pm$ 0.002$\bullet$ & 0.198 $\pm$ 0.002$\bullet$ & 0.033 $\pm$ 0.001$\bullet$ & 0.074 $\pm$ 0.001$\bullet$ & 0.444 $\pm$ 0.004$\bullet$ & 0.023 $\pm$ 0.000 \\
        \hline
        \multirow{4}[2]{*}{Corel}
        &SC$\uparrow$
        & 0.112 $\pm$ 0.014$\circ$ & -0.052 $\pm$ 0.000$\bullet$ & -0.070 $\pm$ 0.000$\bullet$ & -0.277 $\pm$ 0.000$\bullet$ & -0.128 $\pm$ 0.000$\bullet$ & 0.238 $\pm$ 0.002$\circ$ & -0.016 $\pm$ 0.000 \\
        &DI$\uparrow$
        & 0.031 $\pm$ 0.001$\bullet$ & 0.032 $\pm$ 0.000$\bullet$ & 0.020 $\pm$ 0.000$\bullet$ & 0.015 $\pm$ 0.000$\bullet$ & 0.019 $\pm$ 0.000$\bullet$ & 0.032 $\pm$ 0.000$\bullet$ & 0.354 $\pm$ 0.000 \\
        &NMI$\downarrow$
        & 0.643 $\pm$ 0.035$\bullet$ & 0.204 $\pm$ 0.002$\bullet$ & 0.209 $\pm$ 0.002$\bullet$ & 0.092 $\pm$ 0.001$\bullet$ & 0.394 $\pm$ 0.006$\bullet$ & 0.762 $\pm$ 0.018$\bullet$ & 0.070 $\pm$ 0.000 \\
        &JC$\downarrow$
        & 0.219 $\pm$ 0.004$\bullet$ & 0.031 $\pm$ 0.001$\bullet$ & 0.029 $\pm$ 0.001$\bullet$ & 0.013 $\pm$ 0.000 & 0.072 $\pm$ 0.001$\bullet$ & 0.410 $\pm$ 0.013$\bullet$ & 0.010 $\pm$ 0.000 \\
        \hline
        \multirow{4}[2]{*}{Digits}
        &SC$\uparrow$
        & -0.133 $\pm$ 0.022$\bullet$ & -0.001 $\pm$ 0.000$\bullet$ & 0.061 $\pm$ 0.000 & -0.076 $\pm$ 0.000$\bullet$ & -0.117 $\pm$ 0.000$\bullet$ & 0.471 $\pm$ 0.013$\circ$ & 0.064 $\pm$ 0.000 \\
        &DI$\uparrow$
        & 0.016 $\pm$ 0.000$\bullet$ & 0.016 $\pm$ 0.000$\bullet$ & 0.018 $\pm$ 0.001$\bullet$ & 0.008 $\pm$ 0.000$\bullet$ & 0.016 $\pm$ 0.001$\bullet$ & 0.062 $\pm$ 0.000$\bullet$ & 0.087 $\pm$ 0.000\\
        &NMI$\downarrow$
        & 0.078 $\pm$ 0.000$\bullet$ & 0.364 $\pm$ 0.012$\bullet$ & 0.399 $\pm$ 0.008$\bullet$ & 0.011 $\pm$ 0.000 & 0.515 $\pm$ 0.022$\bullet$ & 0.822 $\pm$ 0.028$\bullet$ & 0.008 $\pm$ 0.000 \\
        &JC$\downarrow$
        & 0.076 $\pm$ 0.000$\bullet$ & 0.279 $\pm$ 0.003$\bullet$ & 0.298 $\pm$ 0.000$\bullet$ & 0.053 $\pm$ 0.000 & 0.279 $\pm$ 0.004$\bullet$ & 0.656 $\pm$ 0.015$\bullet$ & 0.052 $\pm$ 0.000 \\
        \hline
        \multirow{4}[2]{*}{Wiki article}
        &SC$\uparrow$
        & 0.447 $\pm$ 0.016$\circ$ & -0.024 $\pm$ 0.000$\bullet$ & -0.031 $\pm$ 0.000$\bullet$ & -0.029 $\pm$ 0.000$\bullet$ & 0.108 $\pm$ 0.002$\circ$ & 0.418 $\pm$ 0.012$\circ$ & 0.066 $\pm$ 0.000 \\
        &DI$\uparrow$
        & 0.124 $\pm$ 0.000 & 0.085 $\pm$ 0.000$\bullet$ & 0.085 $\pm$ 0.001$\bullet$ & 0.083 $\pm$ 0.000$\bullet$ & 0.095 $\pm$ 0.001$\bullet$ & 0.135 $\pm$ 0.001$\circ$ & 0.122 $\pm$ 0.000 \\
        &NMI$\downarrow$
        & 0.803 $\pm$ 0.019$\bullet$ & 0.042 $\pm$ 0.000$\bullet$ & 0.041 $\pm$ 0.000$\bullet$ & 0.006 $\pm$ 0.000 & 0.212 $\pm$ 0.001$\bullet$ & 0.783 $\pm$ 0.052$\bullet$ & 0.006 $\pm$ 0.000 \\
        &JC$\downarrow$
        & 0.593 $\pm$ 0.006$\bullet$ & 0.078 $\pm$ 0.001$\bullet$ & 0.078 $\pm$ 0.000$\bullet$ & 0.056 $\pm$ 0.001 & 0.113 $\pm$ 0.002$\bullet$ & 0.535 $\pm$ 0.014$\bullet$ & 0.052 $\pm$ 0.000 \\
        \hline
        \multirow{4}[2]{*}{Mirflickr}
        &SC$\uparrow$ & -0.004 $\pm$ 0.000$\circ$ & -0.092 $\pm$ 0.000$\bullet$ & -0.028 $\pm$ 0.000$\circ$ & -0.058 $\pm$ 0.000$\bullet$ & -0.093 $\pm$ 0.000$\bullet$ & 0.017 $\pm$ 0.000$\circ$ & -0.038 $\pm$ 0.000 \\
        &DI$\uparrow$ & 0.061 $\pm$ 0.002$\bullet$ & 0.062 $\pm$ 0.005$\bullet$ & 0.071 $\pm$ 0.000$\bullet$ & 0.053 $\pm$ 0.001$\bullet$ & 0.064 $\pm$ 0.001$\bullet$ & 0.059 $\pm$ 0.002$\bullet$ & 0.173 $\pm$ 0.005\\
        &NMI$\downarrow$ & 0.427 $\pm$ 0.012$\bullet$ & 0.016 $\pm$ 0.000$\bullet$ & 0.021 $\pm$ 0.000$\bullet$ & 0.014 $\pm$ 0.000$\bullet$ & 0.216 $\pm$ 0.006$\bullet$ & 0.575 $\pm$ 0.011$\bullet$ & 0.005 $\pm$ 0.000 \\
        &JC$\downarrow$ & 0.878 $\pm$ 0.022$\bullet$ & 0.047 $\pm$ 0.000$\bullet$ & 0.037 $\pm$ 0.000$\bullet$ & 0.023 $\pm$ 0.000 & 0.073 $\pm$ 0.000$\bullet$ & 0.368 $\pm$ 0.011$\bullet$ & 0.022 $\pm$ 0.000 \\
        \hline
\end{tabular}
\label{table2}}
\end{table*}

\begin{table}[t]
\scriptsize
		\caption{Quality and Diversity of the various competing methods on finding multiple co-clusterings. $\bullet / \circ$ indicates whether MVMCC is statistically (according to pairwise $t$-test at 95\% significance level) superior/inferior to the other method.}
		\centering
\begin{tabular}{c| c| c c c}
	\hline
	& & MCC-NBMM &MultiCC &MVMCC\\
    \hline
        \multirow{4}[2]{*}{Caltech-7}
        &SC$\uparrow$
        & -0.100 $\pm$ 0.002$\bullet$ & -0.103 $\pm$ 0.006$\bullet$ & 0.198 $\pm$ 0.004 \\
        &DI$\uparrow$
        & 0.034 $\pm$ 0.000$\bullet$ & 0.011 $\pm$ 0.000$\bullet$ & 0.047 $\pm$ 0.000 \\
        &NMI$\downarrow$
        & 0.376 $\pm$ 0.014$\bullet$ & 0.005 $\pm$ 0.000 & 0.005 $\pm$ 0.000 \\
        &JC$\downarrow$
        & 0.185 $\pm$ 0.003$\bullet$ & 0.087 $\pm$ 0.000 & 0.083 $\pm$ 0.000 \\
    \hline
        \multirow{4}[2]{*}{Caltech-20}
        &SC$\uparrow$
        & -0.134 $\pm$ 0.000$\bullet$ & -0.229 $\pm$ 0.012$\bullet$ & 0.080 $\pm$ 0.000 \\
        &DI$\uparrow$
        & 0.026 $\pm$ 0.000$\bullet$ & 0.011 $\pm$ 0.000$\bullet$ & 0.156 $\pm$ 0.008 \\
        &NMI$\downarrow$
        & 0.325 $\pm$ 0.010$\bullet$ & 0.021 $\pm$ 0.000 & 0.026 $\pm$ 0.000 \\
        &JC$\downarrow$
        & 0.150 $\pm$ 0.002$\bullet$ & 0.056 $\pm$ 0.000$\bullet$ & 0.029 $\pm$ 0.000 \\
    \hline
        \multirow{4}[2]{*}{Corel}
        &SC$\uparrow$
        & -0.087 $\pm$ 0.002$\bullet$ & -0.172 $\pm$ 0.012$\bullet$ & -0.017 $\pm$ 0.000 \\
        &DI$\uparrow$
        & 0.024 $\pm$ 0.000$\bullet$ & 0.015 $\pm$ 0.000$\bullet$ & 0.152 $\pm$ 0.002 \\
        &NMI$\downarrow$
        & 0.377 $\pm$ 0.013$\bullet$ & 0.164 $\pm$ 0.002$\bullet$ & 0.070 $\pm$ 0.000 \\
        &JC$\downarrow$
        & 0.176 $\pm$ 0.004$\bullet$ & 0.044 $\pm$ 0.000$\bullet$ & 0.010 $\pm$ 0.000 \\
    \hline
        \multirow{4}[2]{*}{Digits}
        &SC$\uparrow$
        & -0.243 $\pm$ 0.024$\bullet$ & -0.214 $\pm$ 0.013$\bullet$ & 0.144 $\pm$ 0.002 \\
        &DI$\uparrow$
        & 0.014 $\pm$ 0.000 & 0.003 $\pm$ 0.000$\bullet$ & 0.018 $\pm$ 0.000 \\
        &NMI$\downarrow$
        & 0.286 $\pm$ 0.006$\bullet$ & 0.010 $\pm$ 0.000$\circ$ & 0.207 $\pm$ 0.003 \\
        &JC$\downarrow$
        & 0.166 $\pm$ 0.001$\bullet$ & 0.060 $\pm$ 0.000$\circ$ & 0.115 $\pm$ 0.000 \\
    \hline
        \multirow{4}[2]{*}{Wiki article}
        &SC$\uparrow$
        & -0.0694 $\pm$ 0.000$\bullet$ & -0.058 $\pm$ 0.000$\bullet$ & 0.064 $\pm$ 0.000 \\
        &DI$\uparrow$
        & 0.079 $\pm$ 0.000 & 0.041 $\pm$ 0.001$\bullet$ & 0.078 $\pm$ 0.000 \\
        &NMI$\downarrow$
        & 0.287 $\pm$ 0.005$\bullet$ & 0.007 $\pm$ 0.000 & 0.006 $\pm$ 0.000 \\
        &JC$\downarrow$
        & 0.127 $\pm$ 0.002$\bullet$ & 0.054 $\pm$ 0.000 & 0.052 $\pm$ 0.000 \\
    \hline
        \multirow{4}[2]{*}{Mirflickr}
        &SC$\uparrow$
        & -0.095 $\pm$ 0.000$\bullet$ & -0.194 $\pm$ 0.002$\bullet$ & 0.064 $\pm$ 0.000 \\
        &DI$\uparrow$
        & 0.052 $\pm$ 0.000$\bullet$ & 0.065 $\pm$ 0.000$\bullet$ & 0.151 $\pm$ 0.003 \\
        &NMI$\downarrow$
        & 0.017 $\pm$ 0.000$\bullet$ & 0.081 $\pm$ 0.000$\bullet$ & 0.005 $\pm$ 0.000 \\
        &JC$\downarrow$
        & 0.052 $\pm$ 0.000$\bullet$ & 0.052 $\pm$ 0.000$\bullet$ & 0.022 $\pm$ 0.000 \\
    \hline
\end{tabular}\label{table3}
\end{table}

Table \ref{table2} reports the average results (of ten independent runs) and standard deviations of these comparing methods on exploring two alternative one-way clusterings with $h=2$. From Table \ref{table2}, we can find MVMC often performs better than these comparing methods across different multi-view datasets, which proves the effectiveness of MVMC on exploring alternative clusterings on multi-view data. MVMC always has the best results on the diversity metrics (NMI and JC). This fact suggests it can find two alternative clusterings with high diversity. MVMC occasionally has a lower value on quality metrics (SC and DI) than some of these comparing methods. That is explainable, since it is a widely-recognized dilemma to obtain alternative clusterings with high-diversity and high-quality, and MVMC has a much larger diversity than these comparing methods. Although these comparing methods employ different techniques to explore alternative clusterings in the subspaces or by reducing the redundancy between the clusterings, they almost always lose to MVMC. The  cause is that the concatenated long feature vectors override the intrinsic structures of different views. It also explains why these comparing methods have lower values on diversity metrics (NMI and JC). In practice, because of the long concatenated feature vectors, these comparing methods generally suffer from a long runtime and cannot be applied on multi-view datasets with high-dimensional  feature views. In contrast, our MVMC is rather efficient, it does not need to concatenate features and is directly applicable on each view.

Table \ref{table3} reports the results of MVMCC, MultiCC and MCC-NBMM in exploring multiple co-clusterings, whose number is equal to the number of views $h=m$ (unlike $h=2$ for Table \ref{table2}), because of the feature view heterogeneity. For this evaluation, we report the average quality and diversity values of all pairwise alternative co-clusterings of   $h$ clusterings.
We can see  MVMCC significantly outperforms these two state-of-the-art multiple co-clustering methods across different evaluation metrics and datasets. MultiCC sometimes obtains a better diversity than MVMCC. That is because it directly optimizes the diversity on the sample-cluster and feature-cluster matrices, while our MVMCC indirectly optimizes the diversity and mainly from the sample-cluster matrices.  These results prove the effectiveness of our solution on exploring multiple co-clusterings on multi-view data.

\begin{table}[t]
\vspace{-2mm}
\scriptsize
		\caption{Comparison results with/without the shared information matrix $\mathbf{U}$ for discovering multiple clusterings.}
		\centering
\begin{tabular}{c|c|c|c|c|c}
	\hline
	 & &SC$\uparrow$ &DI$\uparrow$ &NMI$\downarrow$ &JC$\downarrow$\\
	\hline
    \multirow{2}{*}{Digits}&MVMC(nU)&0.061&0.076&0.008&0.050\\
    &MVMC&0.064&0.087&0.008&0.052\\
    \cline{1-6}
	\multirow{2}{*}{Wiki article}&MVMC(nU)&0.059&0.088&0.005&0.051\\
    &MVMC&0.066&0.122&0.006&0.052\\
	\hline
\end{tabular}
\label{table4}
\end{table}

Following the experimental setup in Table \ref{table2}, we conduct additional experiments to investigate the contribution of shared matrix $\mathbf{U}$ on improving the quality of multiple clusterings. For this investigation, we introduce a variant of MVMC(nU), which only uses ${\{{\mathbf{D}^k}\}}_{k=1}^h$ to generate multiple clusterings and disregards the shared information matrix $\mathbf{U}$. From the reported results on  Digits and Wikipeda datasets in Table \ref{table4}, we can find the diversity (NMI and JC) of two alternative clusterings keeps almost the same when the shared information matrix $\mathbf{U}$ is excluded. However, the quality (SC and DI) is clearly reduced when $\mathbf{U}$ is excluded. This contrast proves that $\mathbf{U}$ indeed improves the quality of multiple clusterings, and justifies our motivation to seek $\mathbf{U}$ across views and leverage it with individuality information matrix $\mathbf{D}^v$ for multiple clusterings.

In summary, we can conclude that the diversity information matrix helps to generate diverse clusterings, and the commonality information matrix evacuated from multi-view data can respectively improve the quality of these clusterings. These experimental results also confirm our assumption that the diversity and commonality of multi-view data can be leveraged to generate diverse clusterings with high-quality.

\vspace{-2mm}
\subsection{Parameter analysis}
\begin{figure}[h!tbp]
\centering
\subfigure[DI]{\label{fig2a}
\includegraphics[width=4cm,height=3cm]{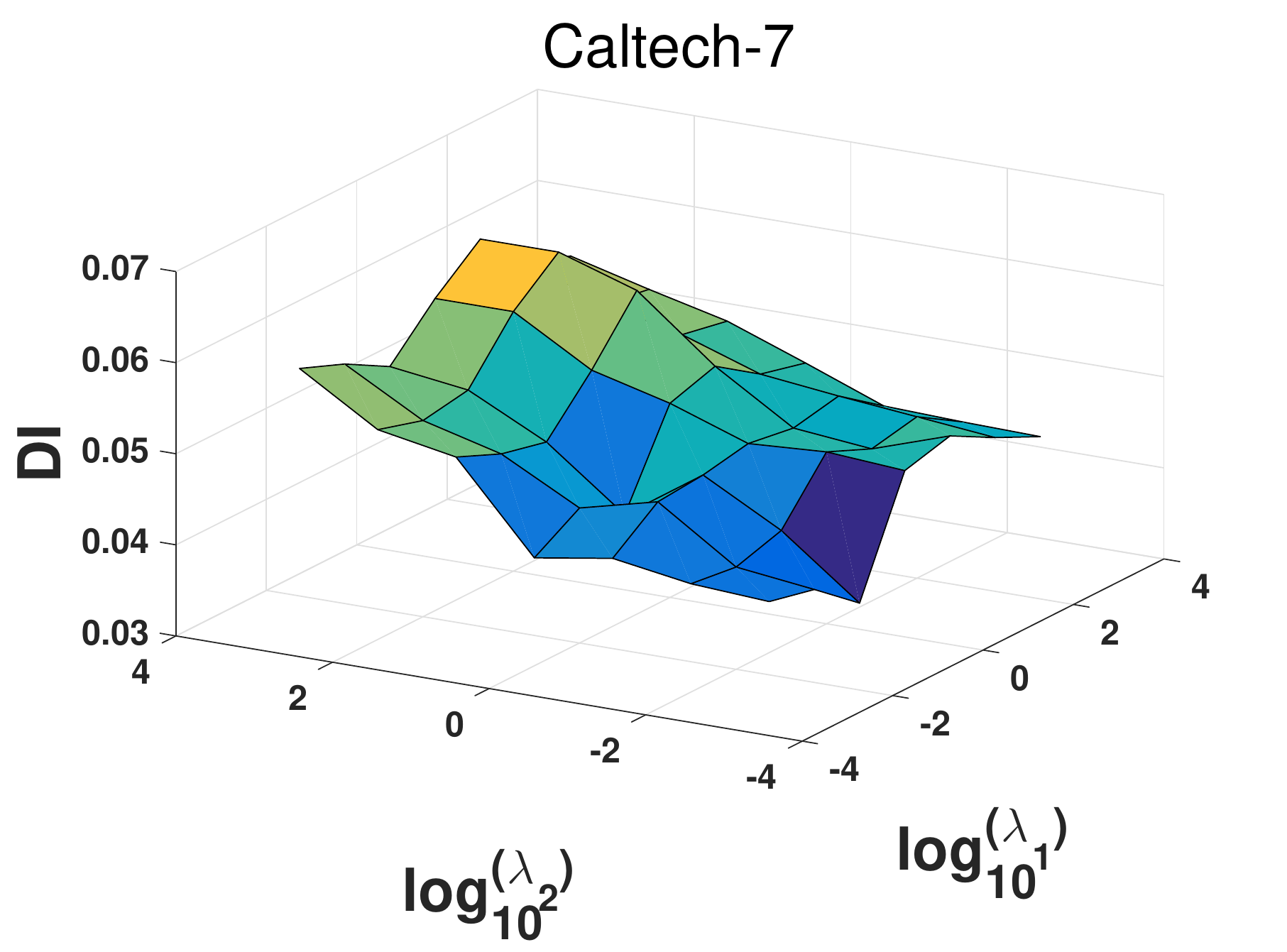}}
\hspace{-1em}
\subfigure[1-NMI]{\label{fig2b}
\includegraphics[width=4cm,height=3cm]{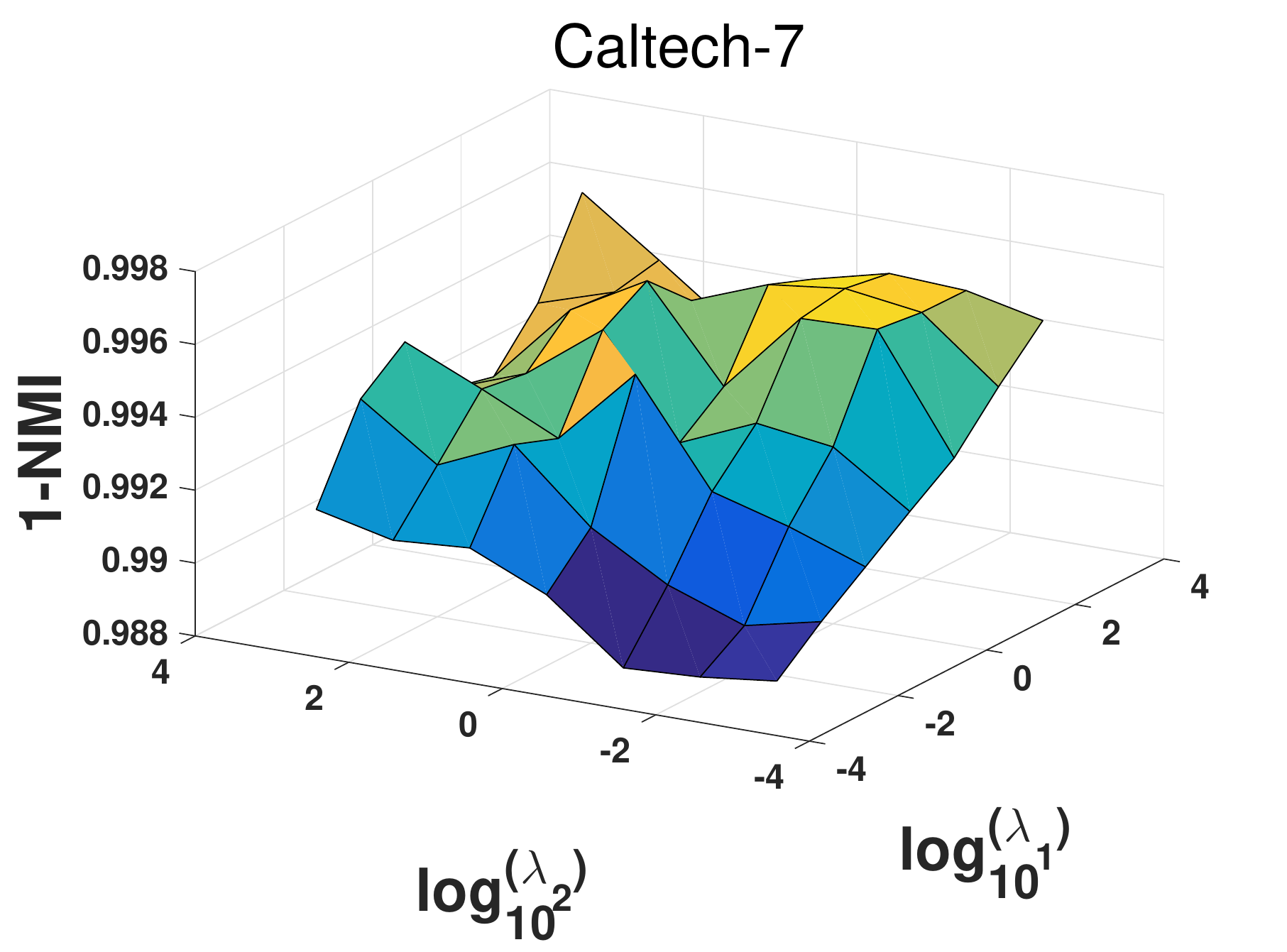}}
\vspace{-1em}
\caption{Quality (SI) and Diversity(1-NMI) of MVMC vs. $\lambda_1$ and $\lambda_2$ on the Caltech-7 dataset.}
\vspace{-1em}
\label{fig2}
\end{figure}

$\lambda_1$ and $\lambda_2$ are two important input parameters of MVMC (MCMCC) for seeking the individuality and commonality information of multi-view data, which consequently affect the quality and diversity of multiple clusterings.
We investigate the sensitivity of MVMC to these parameters by varying $\lambda_1$ (it controls diversity) and $\lambda_2$ (it controls quality) in the range $[10^{-3}, 10^{-2}, \cdots,10^3]$. Figure \ref{fig2} reports the Quality (DI) and Diversity (1-NMI, the larger the better) of MVMC on the Caltech-7 dataset. We have several interesting observations: (i) diversity (1-NMI) increases as  $\lambda_1$ increases, but not  as much as with the increase of $\lambda_2$ (see Figure \ref{fig2b}); quality (DI) increases as  $\lambda_2$ increases, but not as much as with the increase of  $\lambda_1$. (ii) The synchronous increase of $\lambda_1$ and $\lambda_2$ does not necessarily give the highest results in quality and diversity. (iii) When both $\lambda_1$ and $\lambda_2$ are fixed to a small value, both quality and diversity are reduced. This fact suggests that both diversity and commonality information of multi-view data should be used for the exploration of alternative clusterings. These observations again confirm the known dilemma between diversity and quality of multiple clusterings. The values $\lambda_1 = 10$ and $\lambda_2 = 100$ often provide the best balance between quality and diversity.

\begin{figure}[h!tbp]
\centering
\subfigure[DI]{\label{fig3a}
\includegraphics[width=3.6cm,height=2.7cm]{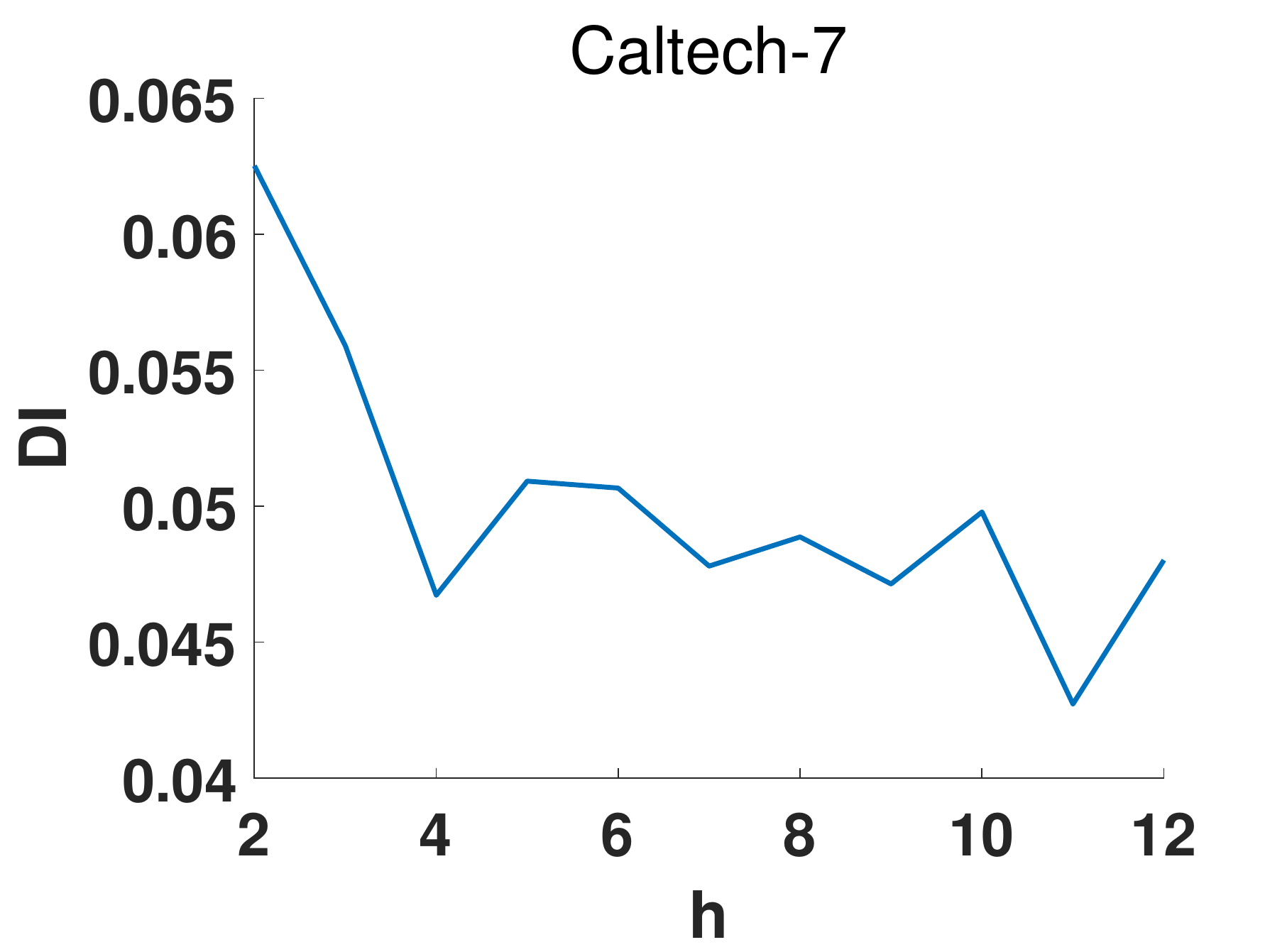}}
\hspace{-1em}
\subfigure[NMI]{\label{fig3b}
\includegraphics[width=3.6cm,height=2.7cm]{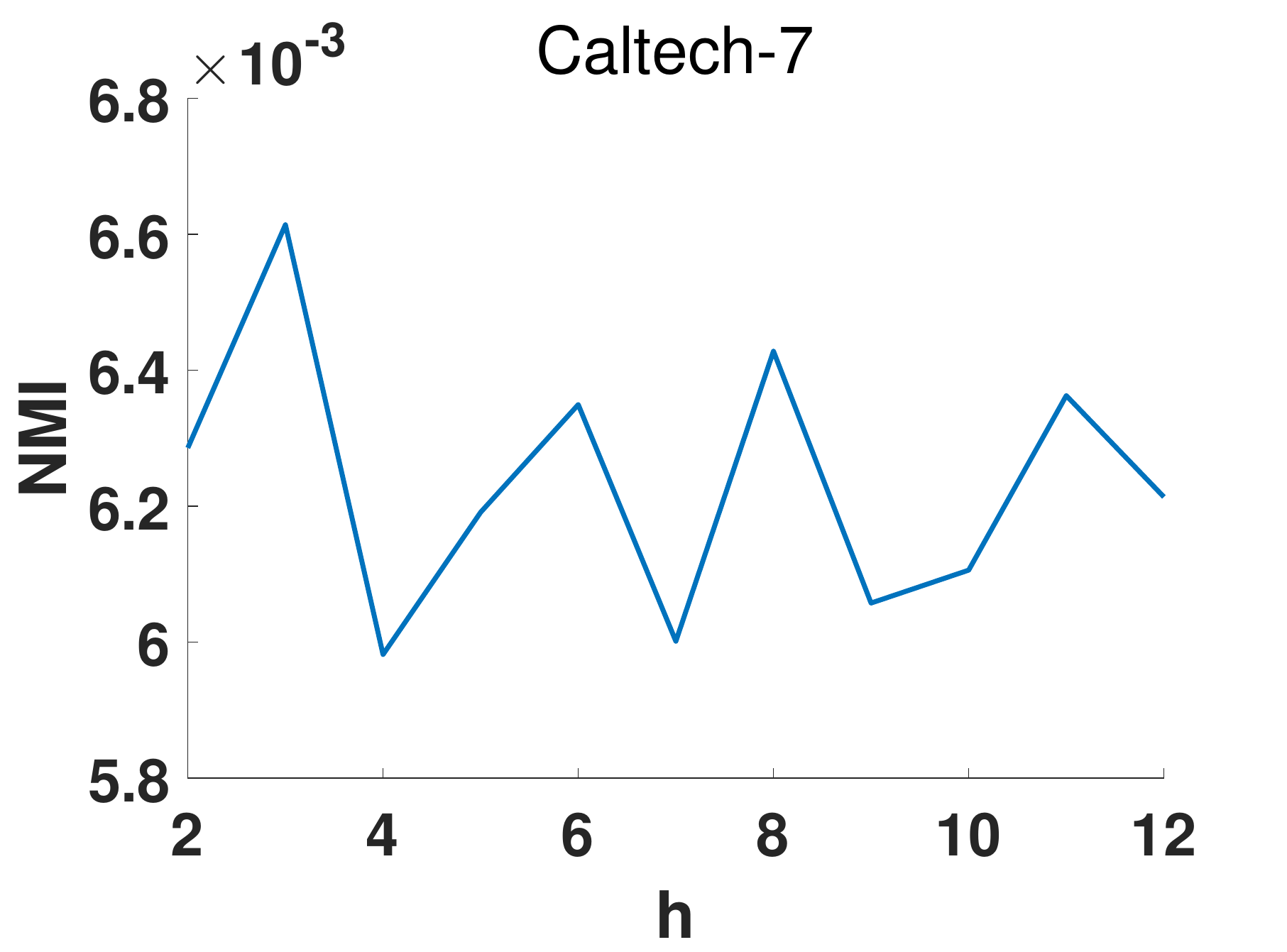}}
\vspace{-1em}
\caption{Quality (SI) and Diversity(NMI, the lower the better) of MVMC vs. $h$ from 2 to $2*m$ on the Caltech-7 dataset.}
\vspace{-1em}
\label{fig3}
\end{figure}

We vary $h$ from 2 to $2*m$ on Caltech-7 dataset to explore the variation of average quality and diversity of multiple clusterings generated by MVMC.  In Figure \ref{fig3}, as the increase of $h$, the average quality of multiple clusterings decreases gradually with small fluctuations, and the average diversity fluctuates in a small range. These patterns are accountable from the dilemma of  quality and diversity of multiple clusterings. More alternative clusterings with diversity scarify the quality of themselves. Overall, we can find that our MVMC can explore $h\geq 2$ alternative clusterings with quality and diversity.

\vspace{-2mm}
\subsection{Runtime Analysis}
Table \ref{table5} gives the runtimes of all methods. The time complexity of our MVMC is $O(t n^2 d (h^2 v + 2hv + 2h))$, where $t$ is the number of iterations for optimization, $v$ is the number of views, $n$ is the number of samples, $d$ is the number of features, $h$ is the number of clusterings. The experiments are conducted on a server with Ubuntu 16.04, Intel Xeon8163 with 1TB RAM; all methods are implemented in Matlab2014a. OSC is the fastest and Dec-kmeans is the second. OSC finds multiple clusterings by iteratively
reducing dimension and applying $k$-means. Dec-kmeans jointly
seeks two different clusterings by minimizing a $k$-means
sum squared error objective for the two clustering solutions,
and the correlation between them. Because $k$-means has a low time complexity, these two $k$-means based methods are much faster than the other techniques.  MVMC and MNMF have similar runtimes, since they are both based on nonnegative matrix factorization, which has a large complexity than $k$-means.  MVMC is more efficient than other comparing methods (except OSC and Dec-kmeans), since it does not need to concatenate features and is directly applicable on each view. In summary, MVMC not only performs better than the state-of-the-art methods in exploring multiple clusterings, but also holds a comparable runtime with the efficient counterparts.

\vspace{-2mm}
\begin{table}[h!tbp]
\scriptsize
		\caption{Runtimes of comparing methods (in seconds).}
		\centering
\resizebox{85mm}{12mm}{
\begin{tabular}{c|c|c|c|c|c|c|c}
	\hline
	& Caltech7 & Caltech20 & Corel & Digits & Wiki & Mirflickr & Total \\
	\hline
    Dec-kmeans & 112 & 273 & 433 & 39 & 14 & 361 & 1232 \\
    ISAAC & 1432 & 2232 & 1822 & 1223 & 656 & 10012 & 17074 \\
    MISC  & 1336 & 2156 & 1956 & 1278 & 638 & 21168 & 28532 \\
    MNMF  & 105 & 1257 & 614 & 356 & 220 & 1577 & 4129 \\
    mSC   & 864 & 1564 & 1269 & 690 & 527 & 2631 & 7545 \\
    OSC   & 4 & 10 & 1 & 1 & 2 & 217 & 235 \\
    MVMC  & 332 & 585 & 790 & 234 & 490 & 9564 & 11995 \\
    \hline
\end{tabular}\label{table5}}
\end{table}

\vspace{-6.5mm}
\section{Conclusion}
In this paper, we proposed an approach to generate multiple clusterings (co-clusterings) from multi-view data, which is an interesting but largely overlooked clustering topic which encompasses multi-view clustering and multiple clusterings. Our approach leverages the diversity and commonality of multi-view data to generate multiple clusterings, and outperforms state-of-the-art multiple clustering solutions. Our study confirms the existence of individuality and commonality of multi-view data, and their contribution for generating diverse clusterings with quality. We plan to find a principal way to automatically determine the number of alternative clusterings for our proposed approach.

\vspace{-3mm}
\bibliographystyle{named}
\bibliography{MVMC}

\end{document}